\documentclass[runningheads]{llncs}
\usepackage{amsmath}
\usepackage{amssymb}
\usepackage{booktabs}
\usepackage{cite}
\usepackage{color}
\usepackage{dirtytalk}
\usepackage{enumitem}
\usepackage{gensymb}
\usepackage{graphicx}
\usepackage{latexsym}
\usepackage{multirow}
\usepackage{nonfloat}
\usepackage[all]{nowidow}
\usepackage{pifont}
\usepackage[caption=false]{subfig}
\usepackage[dvipsnames]{xcolor}
\usepackage{xfrac}
\usepackage[pagebackref=false,breaklinks=true,colorlinks=true,bookmarks=false]{hyperref}

\newcommand{\cmark}{\ding{51}}
\newcommand{\xmark}{\ding{55}}

\begin{document}
%
\title{Attention to Warp: Deep Metric Learning for Multivariate Time Series}
%
%
\author{Shinnosuke Matsuo\inst{1} \and
Xiaomeng Wu\inst{2}\orcidID{0000-0003-2816-1781} \and
Gantugs Atarsaikhan\inst{1} \and
Akisato Kimura\inst{2} \and
Kunio Kashino\inst{2} \and
Brian Kenji Iwana\inst{1}\orcidID{0000-0002-5146-6818} \and
Seiichi Uchida\inst{1}\orcidID{0000-0001-8592-7566}
}
%
\authorrunning{S. Matsuo et al.}
%
\institute{Kyushu University, Fukuoka, Japan
\email{shinnosuke.matsuo@human.ait.kyushu-u.ac.jp} 
\and
Communication Science Laboratories, NTT Corporation, Japan
}

\maketitle              

\setcounter{footnote}{0}


\begin{abstract}
Deep time series metric learning is challenging due to the difficult trade-off between \emph{temporal invariance} to nonlinear distortion and \emph{discriminative power} in identifying non-matching sequences. This paper proposes a novel neural network-based approach for robust yet discriminative time series classification and verification. This approach adapts a parameterized attention model to time warping for greater and more adaptive \emph{temporal invariance}. It is robust against not only local but also large global distortions, so that even matching pairs that do not satisfy the monotonicity, continuity, and boundary conditions can still be successfully identified. Learning of this model is further guided by dynamic time warping to impose temporal constraints for stabilized training and higher \emph{discriminative power}. It can learn to augment the inter-class variation through warping, so that similar but different classes can be effectively distinguished. We experimentally demonstrate the superiority of the proposed approach over previous non-parametric and deep models by combining it with a deep online signature verification framework, after confirming its promising behavior in single-letter handwriting classification on the Unipen dataset.


\keywords{Attention model \and Dynamic time warping \and Metric learning \and Signature verification.}
\end{abstract}

\section{Introduction}

Over the past two decades, time series classification and verification have been considered to be two of the most challenging problems in data mining. They have been applied to many applications, such as activity recognition, computational auditory scene analysis, cybersecurity, electronic health records, and handwritten biometric recognition~\cite{FawazFWIM19}. The fundamental problem in these tasks is to define the distance between time series. This distance must be invariant to nonlinear temporal distortions, e.g., minute but diverse shifts, scaling, and noise, due to intra-class variations. It should also be able to capture temporal inconsistency so that time series of different classes can be optimally distinguished.

\begin{figure}[t]
\centering
\hspace*{\fill}
\subfloat[DTW]{
\includegraphics[scale=.45]{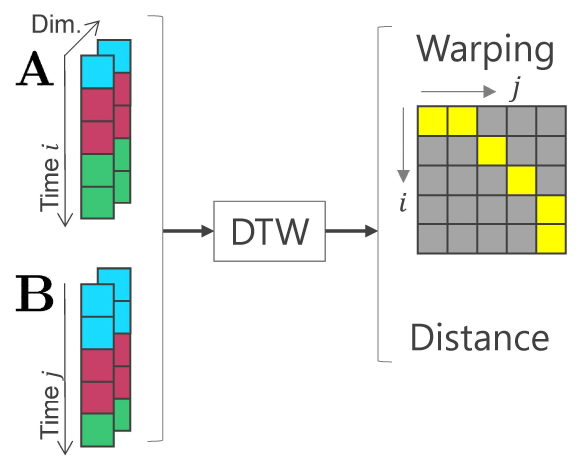}
\label{subf:purpose-dtw}}
\hfill
\subfloat[Proposed Approach]{
\includegraphics[scale=.45]{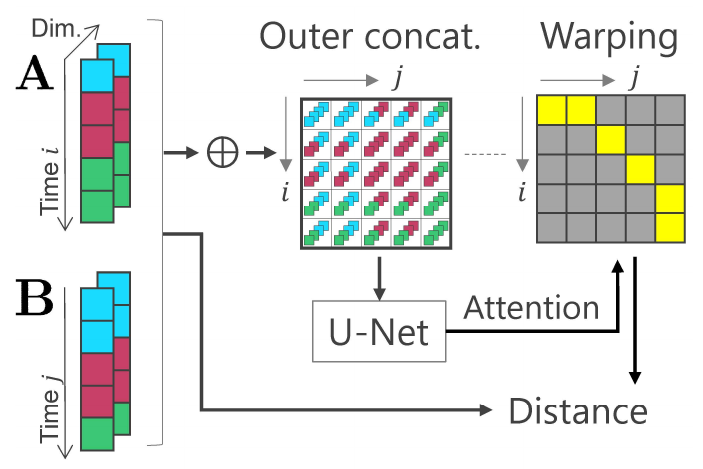}
\label{subf:purpose-prop}}
\hspace*{\fill}
\caption{Comparison between DTW and our approach. While DTW creates the warping path and distance with a non-parametric algorithm, our approach learns warping through an attention model where the attention (weights) form the warping path.}
\label{f:purpose}
\end{figure}

Dynamic Time Warping (DTW)~\cite{DTW78} is the most widely used measure for this purpose~\cite{KholmatovY05,Marteau09,RakthanmanonCMBWZZK12,LinesB15,BagnallLBLK17}. It calculates the minimum temporal alignment cost between two time series by taking the summation of the distances between local elements that are optimally matched under monotonicity, continuity, and boundary conditions (Fig.~\ref{subf:purpose-dtw}). By defining this minimum cost as a dissimilarity measure, DTW achieves invariance to temporal distortions and imposes strong constraints on the measure. However, DTW relies on handcrafted features and may be less effective when complex feature representations are required. Meanwhile, Hidden Markov Models (HMMs)~\cite{GeS00,Fierrez-AguilarORG07} are often reported to be well-suited for time series classification thanks to their high adaptability to intra-class temporal variations. Similar to DTW, their limited complexity makes them difficult to take advantage of the information in the training data.



In recent years, more and more efforts have been made on neural network-based classification and metric learning~\cite{BromleyGLSS93,MuellerT16,CheHXL17,CoskunTCNT18,GrabockaS18,RoyMM18,TolosanaVFO18,AhrabianB19,WuKUK19,WuKIUK19} for discrete time series. In deep metric learning, a Siamese network or a triplet network is used to learn the distance metric from data, driving the distance to be small for \emph{matching pairs} (data from the same class) and large for \emph{non-matching pairs} (data from different classes). Some networks~\cite{MuellerT16,CoskunTCNT18,RoyMM18} produce a global feature representation for the time series, inflicting the heavy loss of useful temporal information. Others~\cite{BromleyGLSS93,TolosanaVFO18,AhrabianB19} encode the time series into a sequence of multidimensional feature vectors that are not explicitly invariant to nonlinear temporal distortions. Tallec and Ollivier~\cite{TallecO18a} theoretically proved that the learnable gates in RNNs formally provide quasi-invariance to temporal transformations. Meanwhile, DECADE~\cite{CheHXL17} and NeuralWarp~\cite{GrabockaS18} were proposed, in which the time series metric is defined as the (weighted) average of all the pairwise distances between deep feature sequences. These methods~\cite{CheHXL17,GrabockaS18,TallecO18a} offer robustness regarding temporal distortions but impose little constraints on the discrimination of non-matching time series. Wu et al.~\cite{WuKUK19,WuKIUK19} proposed to incorporate the DTW into a Siamese network for simultaneous temporal invariance and discriminative power. However, these two approaches remain inefficient because DTW is a sequential algorithm and, unlike convolutional and fully connected layers, it is hard to take advantage of parallel GPU computing.


In this paper, we propose a novel deep model for time series metric learning. The model should be robust with regards to \emph{temporal invariance} to distortion. Meanwhile, it should be able to impose temporal constraints on the learned metric for higher \emph{discriminative power} in terms of non-matching pairs. To this end, we first propose an adaptation of a parameterized attention model to metric learning for greater and more adaptive temporal invariance. Given two time series, which can be raw signals or feature sequences extracted from a Siamese network, the attention model predicts a flexible decision for each pair of temporal locations regarding how proper it is to align the local data points at the two locations in time. The rectangular array of these decisions forms a warping matrix and is used for nonlinear distortion rectification (Fig.~\ref{subf:purpose-prop}). This proposal is inspired by the great success of attention models employed in image captioning and machine translation~\cite{BahdanauCB14,LuongPM15,VaswaniSPUJGKP17}. Metric learning drives the distance between a time series and a warped counterpart to be small for matching pairs and large for non-matching pairs.


We further propose a novel pre-training strategy to initialize the weights of the proposed model more effectively. The pre-training is guided by DTW so that temporal constraints can be imposed on the warping matrix for higher discriminative power. By applying the proposed approach to single-letter handwriting classification and online signature verification, we experimentally demonstrate our superior performance over conventional non-parametric and deep models for time series metric learning.


\section{Proposed Approach}
\label{s:proposed}

\subsection{Overview}

\begin{figure}[t]
\centering
\includegraphics[width=.7\linewidth]{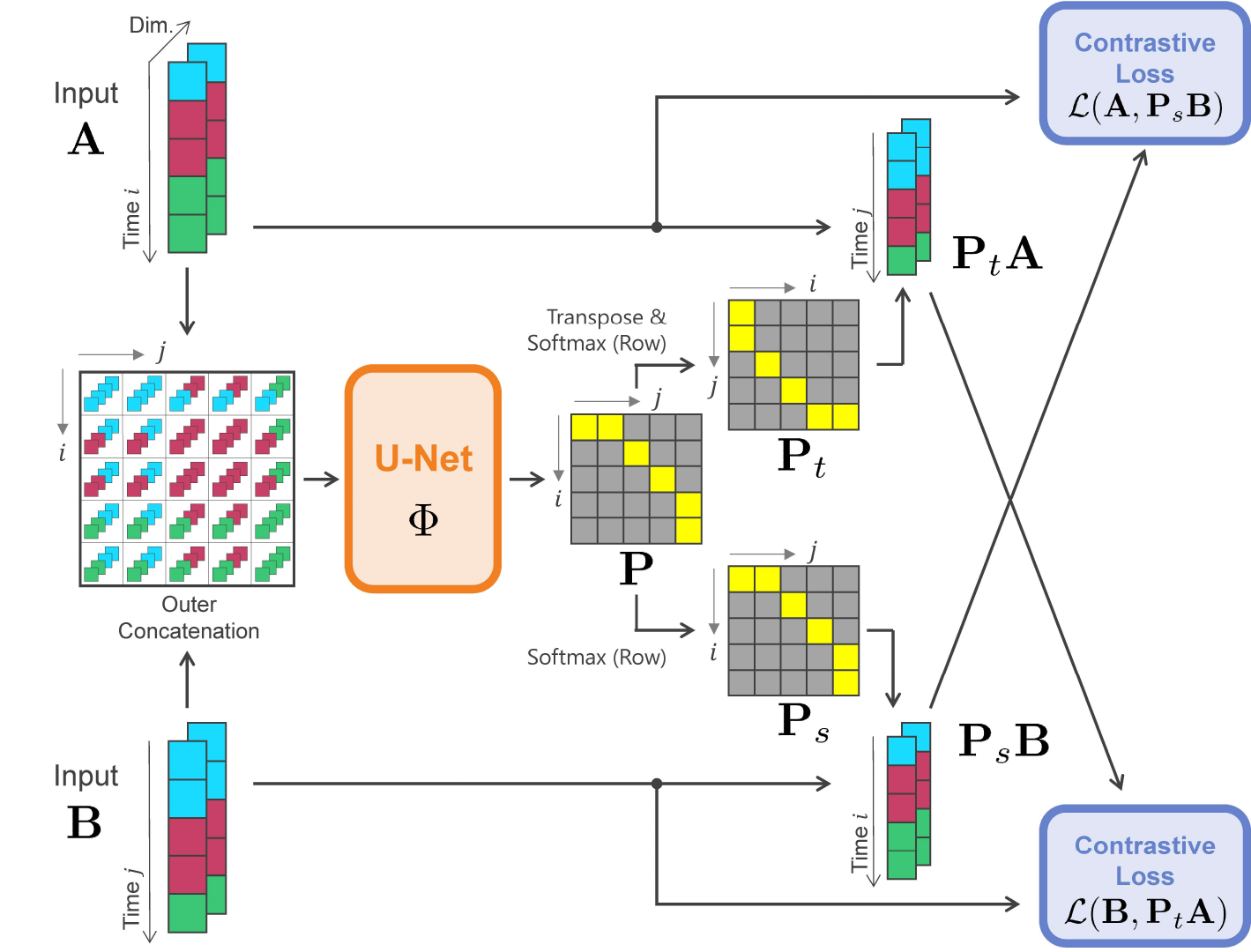}
\caption{Overview of our proposed network (training stage).}
\label{f:overview}
\end{figure}

A schematic overview of the proposed network is shown in Fig.~\ref{f:overview}. Given two time series $\mathbf{A}$ and $\mathbf{B}$ as inputs, our network aims to output a distance that measures their dissimilarity. The time series can be raw signal sequences or deep feature sequences extracted from a Siamese network. The two inputs are first reshaped and concatenated to form an \say{outer concatenation} (see Section~\ref{subs:attention} for details), which is further passed to a fully convolutional network (FCN), i.e., the U-net in Fig.~\ref{f:overview}. The output of the FCN is a matrix $\mathbf{P}$ composed of flexible decisions indicating which local data points in $\mathbf{A}$ and $\mathbf{B}$ should be aligned in time. Row-wise softmax is carried out to generate two \say{warping paths} $\mathbf{P}_s$ and $\mathbf{P}_t$, which are used to warp $\mathbf{B}$ and $\mathbf{A}$, respectively. During testing, the time series distance is thus measured based on the inputs and their warped counterparts. In this network, the combination of outer concatenation, FCN, and row-wise softmax corresponds to the attention model adapted to time series metric learning.

We want the distance measure described above to be small for matching pairs and large for non-matching pairs during training. To this end, we define a contrastive loss~\cite{HadsellCL06} based on the distance, as shown in Fig.~\ref{f:overview} and Section~\ref{subs:attention}. We initialize the weights of the network through pre-training, which is guided by DTW for improved discriminative power and is described in Section~\ref{subs:pre-training}.



\subsection{Attention to Warp}
\label{subs:attention}

In the family of attention models, additive attention~\cite{BahdanauCB14} and (scaled) dot-product attention~\cite{LuongPM15,VaswaniSPUJGKP17} are the most widely used. In this study, we choose the parameterized additive attention mechanism and adapt it for time warping to ensure that the learned metric is explicitly invariant to temporal distortions.

Let $\mathbf{A}\in\mathbb{R}^{W\times K}$ and $\mathbf{B}\in\mathbb{R}^{W\times K}$ denote two multivariate time series, where $W$ is the temporal length and $K$ is the number of dimensions (variables). The time series can be a sequence of raw signals, e.g., an online signature, or a sequence of feature vectors, e.g., a deep feature sequence extracted from a Siamese network. We consider a function $\mathrm{\Phi}:\mathbb{R}^{W\times K}\times\mathbb{R}^{W\times K}\mapsto\mathbb{R}^{W\times W}$ that produces a warping matrix $\mathbf{P}=\mathrm{\Phi}(\mathbf{A},\mathbf{B})$ such that $\mathbf{B}$ can be aligned in time with $\mathbf{A}$ in the form of $\mathbf{PB}$. Let $\mathbf{p}_i\in\mathbb{R}^{1\times W}$ be the $i$-th row vector of $\mathbf{P}$ with $i\in[1,W]$. Suppose that $\mathbf{p}_i$ has been $l^1$-normalized. The warped time series $\mathbf{PB}$ can thus be understood as a sequence of $K$-dimensional vectors $\{\mathbf{p}_i\mathbf{B}\}$, each corresponding to a weighted average of all the $K$-dimensional vectors in $\mathbf{B}$. Based on this warping, we expect that each $\mathbf{p}_i\mathbf{B}$ best matches with the $K$-dimensional vector at the $i$-th time step of $\mathbf{A}$, so that intra-class temporal distortions can be accommodated.

The function $\mathrm{\Phi}$ can be decomposed into $W\times W$ kernel functions denoted by $\varphi$. Let $\mathbf{a}_i\in\mathbf{R}^{1\times K}$ and $\mathbf{b}_j\in\mathbf{R}^{1\times K}$ be the vectors of $\mathbf{A}$ and $\mathbf{B}$ at the $i$-th and $j$-th time steps, respectively. Let $p_{ij}$ be the $j$-th element in $\mathbf{p}_i$. The kernel function $p_{ij}=\varphi(\mathbf{a}_i,\mathbf{b}_j)$ thus transforms $\mathbf{a}_i$ and $\mathbf{b}_j$ into a flexible decision regarding whether the two vectors best match each other ($p_{ij}\approx1$) or not ($p_{ij}\approx0$). According to additive attention~\cite{BahdanauCB14}, $\varphi$ can be defined by a neural network consisting of one or more fully connected layers, which takes the concatenation of $\mathbf{a}_i$ and $\mathbf{b}_j$ as the input.

In the case of the proposed approach, the attention model can be realized more efficiently. Let $\mathbf{A}$ be reshaped to a $W\times1\times K$ tensor and be \emph{horizontally} stacked $W$ times. Similarly, $\mathbf{B}$ is reshaped to a $1\times W\times K$ tensor and is \emph{vertically} stacked $W$ times. They are then concatenated along their third axis, leading to a $W\times W\times 2K$ tensor. We call this operation an \say{outer concatenation} (analogous to the outer product) and denote it by $\mathbf{A}\oplus\mathbf{B}$. Instead of obtaining each $p_{ij}$ individually, we can obtain the warping matrix $\mathbf{P}$ at once by the convolution between the outer concatenation $\mathbf{A}\oplus\mathbf{B}$ and the kernel function $\varphi$. Our attention model $\mathrm{\Phi}$ can thus be defined by an FCN.

\noindent
\textbf{Backbone Network.} In this paper, we adopt U-net~\cite{RonnebergerFB15}, which was originally developed for biomedical image segmentation, as the FCN to define $\mathrm{\Phi}$. Since the input (outer concatenation) $\mathbf{A}\oplus\mathbf{B}$ of the FCN can be regarded as a $2K$-channel \say{image} of size $W\times W$, and the output (warping matrix) $\mathbf{P}$ as a single-channel \say{image} of the same size, it is appropriate to employ U-net as the FCN for direct \say{image-to-image} mapping. Compared to the $1\times1$ convolution described above, the U-net linearly increases the size of the receptive field by stacking multiple $3\times3$ convolution layers. The flexible decision $p_{ij}$ thus depends not only on the local data points $\mathbf{a}_i$ and $\mathbf{b}_j$ but also on their neighborhood along the time axis. This enables the exploitation of beneficial contextual information. {Note that for simplicity, we have assumed that the input time series $\mathbf{A}$ and $\mathbf{B}$ have the same length $W$, but our approach is not limited to this assumption since U-net can naturally handle inputs (outer concatenation $\mathbf{A}\oplus\mathbf{B}$) of arbitrary size.}

\noindent\textbf{Learning.} Once the warping matrix $\mathbf{P}$ is output from the U-net, a row-wise softmax is conducted on $\mathbf{P}$ to ensure that each $\mathbf{p}_i\in\mathbf{P}$ is $l^1$-normalized and to make it peakier. $\mathbf{B}$ is then warped according to $\mathbf{P}_s\mathbf{B}$ to compensate for temporal distortions, where $\mathbf{P}_s$ is the output of the softmax. During training, a contrastive loss can thus be defined by Eq.~\eqref{e:contrastive_A}, which is adapted from the extensively used conventional contrastive loss~\cite{HadsellCL06}.

\begin{equation}
\mathcal{L}(\mathbf{A}, \mathbf{P}_s\mathbf{B})=
\begin{cases}
 \frac{1}{WK} \|\mathbf{A}-\mathbf{P}_s\mathbf{B}\|_\mathrm{F}^2&\text{if $z=1$}\\
\max\left(0,\tau- \frac{1}{WK} \|\mathbf{A}-\mathbf{P}_s\mathbf{B}\|_\mathrm{F}^2\right)&\text{otherwise}
\end{cases}
\label{e:contrastive_A}
\end{equation}

\noindent
Here, $\tau$ is the margin of the hinge loss, and $z\in\{0,1\}$ defines whether $\mathbf{A}$ and $\mathbf{B}$ are non-matching or matching, respectively. The distance between the two time series is defined by the Frobenius norm of the difference matrix $\mathbf{A}-\mathbf{P}_s\mathbf{B}$. Minimizing Eq.~\eqref{e:contrastive_A} drives this distance to be small for matching pairs and large for non-matching pairs.

To make the model symmetric, another warping path $\mathbf{P}_t$ is created by transposing $\mathbf{P}$ and then executing row-wise softmax. $\mathbf{A}$ is warped according to $\mathbf{P}_t\mathbf{A}$. A second contrastive loss can thus be defined between $\mathbf{B}$ and $\mathbf{P}_t\mathbf{A}$ as in Eq.~\eqref{e:contrastive_B}.

\begin{equation}
\mathcal{L}(\mathbf{B}, \mathbf{P}_t\mathbf{A})=
\begin{cases}
 \frac{1}{WK} \|\mathbf{B}-\mathbf{P}_t\mathbf{A}\|_\mathrm{F}^2&\text{if $z=1$}\\
\max\left(0,\tau- \frac{1}{WK} \|\mathbf{B}-\mathbf{P}_t\mathbf{A}\|_\mathrm{F}^2\right)&\text{otherwise}
\end{cases}
\label{e:contrastive_B}
\end{equation}

\noindent
\textbf{Inference.} During testing, the proposed model takes two times series as inputs and predicts a distance measuring their dissimilarity. This distance can be applied to a $k$-NN classifier for classification or a threshold classifier for verification. Specifically, we define the distance by Eq.~\eqref{e:distance_proposed}, which is the average of the distances between $\mathbf{A}$ and $\mathbf{P}_s\mathbf{B}$ and between $\mathbf{B}$ and $\mathbf{P}_t\mathbf{A}$.

\begin{equation}
d(\mathbf{A}, \mathbf{B}, \mathbf{P})=
 \frac{1}{2WK} \left(\|\mathbf{A}-\mathbf{P}_s\mathbf{B}\|_\mathrm{F}^2 + \|\mathbf{B}-\mathbf{P}_t\mathbf{A}\|_\mathrm{F}^2\right)
\label{e:distance_proposed}
\end{equation}

\subsection{Pre-training Guided by Dynamic Time Warping}
\label{subs:pre-training}

In our experiments, we found that the widely-used He initialization~\cite{HeZRS15} could not initialize the weights of the proposed model effectively on some datasets. Specifically, the contrastive losses in Eqs.~\eqref{e:contrastive_A} and \eqref{e:contrastive_B} do not decrease as the training stage progresses. The reason might be because the attention model takes no temporal constraint into account and is limited for distinguishing non-matching pairs of time series.



\begin{figure}[t]
\centering
\includegraphics[width=.7\linewidth]{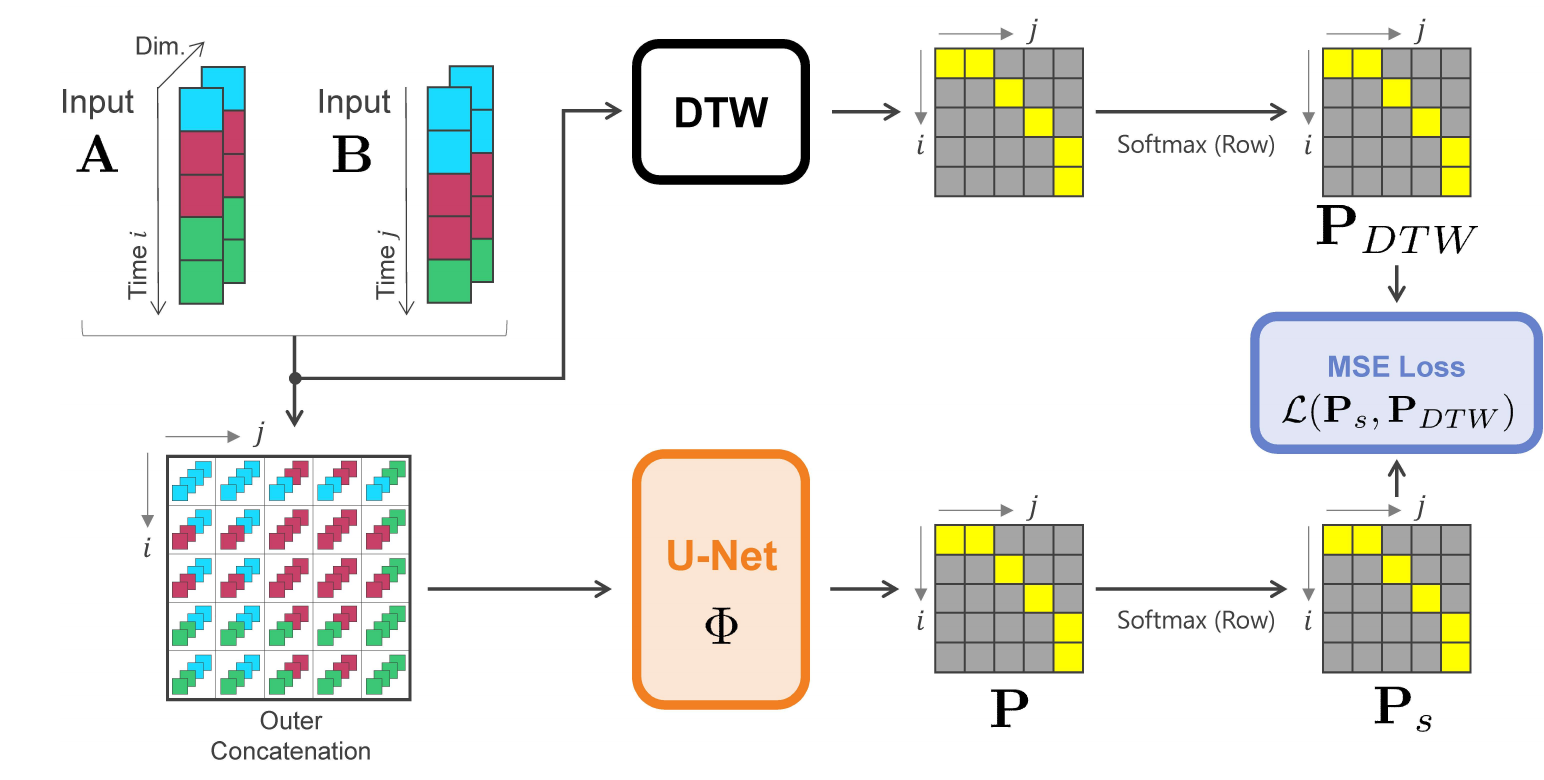}
\caption{Overview of our proposed network (pre-training stage).}
\label{f:overview_pre-training}
\end{figure}


In this paper, we incorporate temporal constraints in our model through pre-training guided by DTW.
Specifically, we pre-train the U-net weights to be used in the proposed model with the guidance of DTW. As shown in Fig.~\ref{f:overview_pre-training}, the input (outer concatenation $\mathbf{A}\oplus\mathbf{B}$ of the two time series) and output (matrix $\mathbf{P}$) of the U-net as well as the row-wise softmax remains the same as those shown in Fig.~\ref{f:overview} and Section~\ref{subs:attention}. The difference from the training stage lies in the loss function. To define the loss, we first apply DTW to the inputs $\mathbf{A}$ and $\mathbf{B}$ and obtain its warping path in the form of a binary matrix. A row-wise softmax is then applied to this matrix, which is similar to the same softmax for $\mathbf{P}$. Let $\mathbf{P}_\mathrm{DTW}$ denote the output of the softmax. Here, we want to make the attention model mimic DTW and force it to produce a warp matrix $\mathbf{P}_s$ that is as close to $\mathbf{P}_\mathrm{DTW}$ as possible. Therefore, we adopt the mean squared error (MSE) between the two matrices as the loss function of pre-training, as shown in Eq.~\eqref{e:pre-training}.

\begin{equation}
\mathcal{L}(\mathbf{P}_s, \mathbf{P}_\mathrm{DTW})=
 \frac{1}{W^2} \|\mathbf{P}_{s}-\mathbf{P}_\mathrm{DTW}\|_\mathrm{F}^2
\label{e:pre-training}
\end{equation}

\noindent
Another loss can also be defined based on the nearly symmetric warping matrix $\mathbf{P}_t$. In this paper, we do not do this for simplicity.

{Note that mimicking DTW during pre-training is to guide the learning of our attention model so that its discriminative power can be improved. Our final purpose is not to mimic DTW but to learn warping through metric learning.}

\section{Experiments}

To evaluate the performance of the proposed approach, we conducted experiments using Unipen~\cite{GuyonSPLJ94}, an online single-letter handwriting dataset, and MCYT-100~\cite{MCYT03}, an online signature dataset.

\subsection{Online Single-Letter Handwriting Classification (Unipen)}
\label{subs:unipen}

Among the multiple subsets officially offered by the provider, Unipen 1a (numerical digits), Unipen 1b (uppercase alphabet), and Unipen 1c (lowercase alphabet) were selected for evaluation. For all three subsets, each time series is a sequence of 2D coordinates of the pen tip, resized to the same fixed temporal length 50. Some statistics are shown in Table~\ref{t:unipen}.

For each class, two sets of 200 data were randomly sampled and used as validation and test data, respectively. The rest of the data were used as training data. When training the network, matching and non-matching pairs were created between all training data. Given a test time series, its distances from all training time series were calculated and a $k$-nearest neighbor ($k$-NN) classifier was used to determine its class label ($k=3$ in this paper).

\smallskip\noindent
\textbf{Baselines.} The proposed approach was compared with the four subsequent baselines. In addition, the performance of the proposed approach without pre-training was also included in the comparison.

\smallskip\noindent
\emph{DTW.} Given a test time series, its distances from the training time series are measured by applying DTW to their raw signals (2D coordinates). This baseline serves as a representative of a non-parametric (non-training) distance measure.

\smallskip\noindent
\emph{Support Vector Machine (SVM).} For this baseline, we trained an SVM with a radial basis function (RBF) kernel for classification. The raw signals were used for training and testing.


\smallskip\noindent
\emph{Classification Network (CN).} This is a convolutional neural network (CNN) that consists of eight convolutional layers and three fully connected (FC) layers. It was trained using a cross-entropy loss.

\smallskip\noindent
\emph{Siamese Network (SN).} This network has almost the same architecture as CN, but was trained with contrastive loss. Its last FC layer produces a global feature representation rather than a probability distribution of classes. Similar to DTW and the proposed approach, SN uses a $k$-NN classifier for classification.

\smallskip\noindent
\textbf{Implementation Details.} The U-Net used in our model consists of an encoder and a decoder that are almost symmetrical, both containing seven convolution layers. The feature maps output by the encoder are concatenated with the corresponding layers of the decoder by skip connections. {Further details can be found in the supplementary document\footnote{\url{http://human.ait.kyushu-u.ac.jp/~matsuo/ICDAR2021_Appendix.pdf}}.} Adam~\cite{ADAM15} was used as the optimizer to train the model, and the learning rate was set to 0.0001. The batch size was set to 512. The margin $\tau$ used in Eqs.~\eqref{e:contrastive_A} and \eqref{e:contrastive_B} was set to one. The training was conducted for up to 20 epochs, and the best model was selected based on the performance measured by the validation data.

\begin{table}[t]
\caption{Classification accuracy (\%) for Unipen.}
\label{t:unipen}
\begin{tabular*}{\linewidth}{@{\extracolsep{\fill}}lccc}
\toprule
~Method & Unipen 1a & Unipen 1b & Unipen 1c \\
\midrule
~Proposed & \textbf{99.0} & \textbf{98.0} & \textbf{95.5} \\
~~~w/o pre-training & 98.4 & 97.3 & 90.7\\
\midrule
~DTW & 98.4 & 96.0 & 94.1 \\
~SVM & 98.2 & 93.9 & 94.1 \\
~CN & 98.3 & 96.2 & \textbf{95.5} \\ 
~SN & 98.6 & 97.2 & \textbf{95.5} \\ 
\midrule
~\#class & 10 & 26 & 26 \\
~\#training (per class) & 756.2 & 232.3 & 391.2 \\
~\#validation (per class) & 200 & 200 & 200 \\
~\#test (per class) & 200 & 200 & 200 \\
\bottomrule
\end{tabular*}
\vspace{2mm}
\end{table}

\smallskip\noindent
\textbf{Results.} The classification results when using Unipen are shown in Table~\ref{t:unipen}. In all three subsets, the proposed approach outperformed or at least was comparable to all baselines. These results confirm the superiority of our deep model in terms of learnable time warping (compared to DTW) and high discriminative power (compared to SVM, CN, and SN). Although the improvement in accuracy is consistent, the range of improvement is not very large. This may be due to a large amount of training data and the short length and low dimensionality of the time series, which render this handwriting classification task a relatively easy classification problem. This also explains the small difference in accuracy between our models w/ and w/o pre-training. In Section~\ref{subs:mcyt}, we apply our model to a more complex verification problem for further evaluation.

\begin{table}[t]
\caption{Number of misclassifications between similar but different classes.}
\label{t:unipen1b_error}
\begin{tabular*}{\linewidth}{@{\extracolsep{\fill}}lccccccc}
\toprule
& \multicolumn{2}{c}{Unipen 1b} & \multicolumn{5}{c}{Unipen 1c} \\
\cmidrule(lr){2-3}
\cmidrule(lr){4-8}
~Method & \texttt{J} vs. \texttt{T} & \texttt{U} vs. \texttt{V} & \texttt{f} vs. \texttt{t} & \texttt{g} vs. \texttt{y} & \texttt{h} vs. \texttt{k} & \texttt{h} vs. \texttt{n} & \texttt{v} vs. \texttt{w} \\
\midrule
~Proposed & \textbf{7} & \textbf{13} & \textbf{10} & \textbf{6} & \textbf{6} & \textbf{13} & \textbf{2} \\
~DTW & 28 & 23 & 16 & 24 & 16 & 20 & 7 \\
\bottomrule
\end{tabular*}
\vspace{2mm}
\end{table}


\begin{figure}[t]
\centering
\subfloat[Unipen 1a]{
\includegraphics[scale=.26]{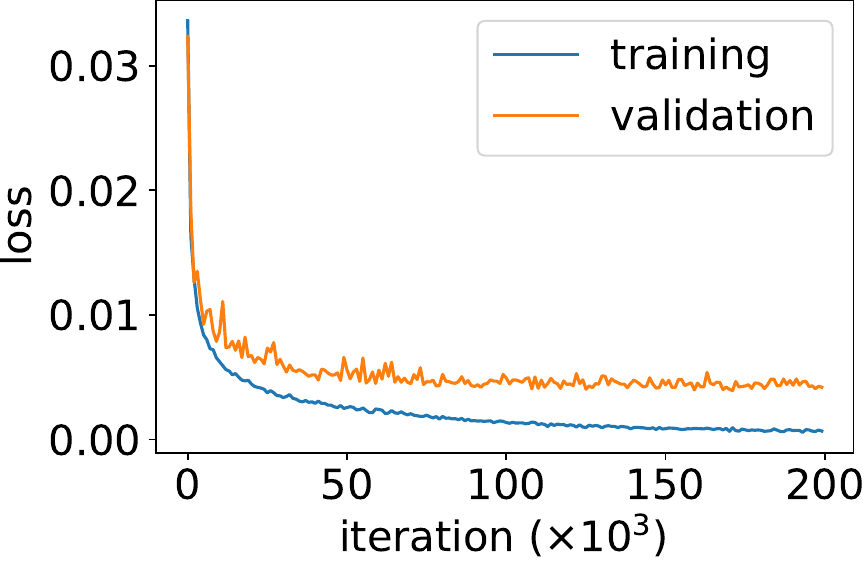}
\label{subf:loss-unipen-1a}}
\hfill
\subfloat[Unipen 1b]{
\includegraphics[scale=.26]{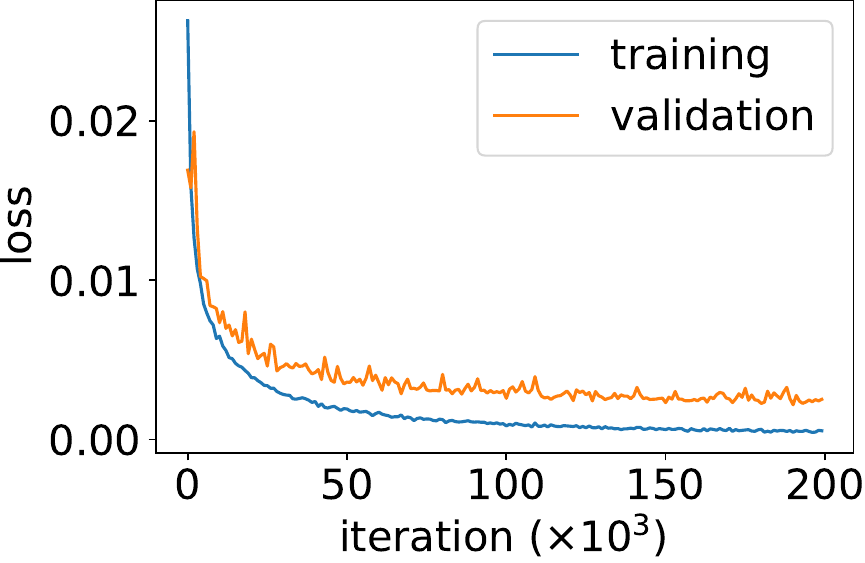}
\label{subf:loss-unipen-1b}}
\hfill
\subfloat[Unipen 1c]{
\includegraphics[scale=.26]{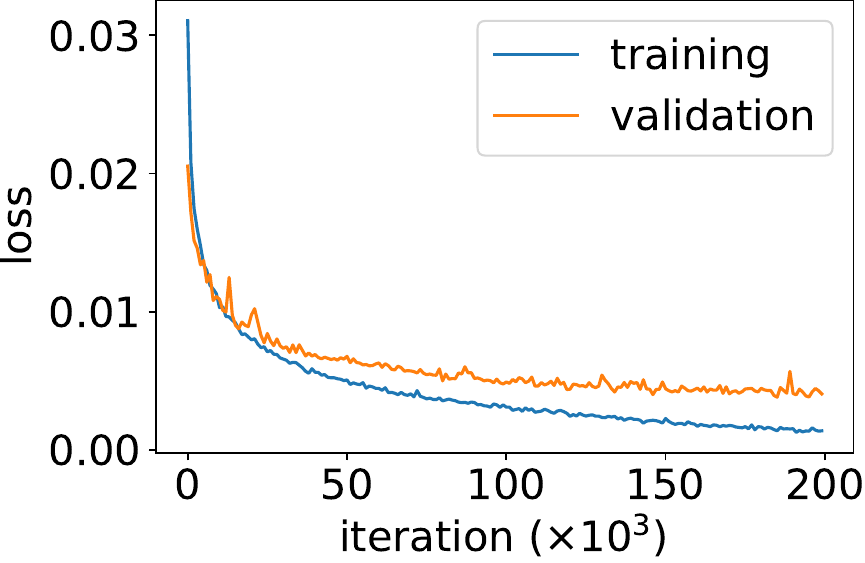}
\label{subf:loss-unipen-1c}}
\caption{Training and validation losses.}
\label{f:loss-unipen}
\end{figure}

Examining the results of Unipen 1b, we found that the largest improvement achieved by the proposed approach lies in the discrimination between \texttt{J} (with a horizontal serif) and \texttt{T} and between \texttt{U} and \texttt{V}. Table~\ref{t:unipen1b_error} compares DTW and the proposed approach in terms of the number of misclassifications between such similar but different alphabetic classes. {Figure~\ref{f:loss-unipen} shows the graphs of training and validation losses, which show good convergence and few signs of overfitting.}

\subsection{Online Signature Verification (MCYT-100)}
\label{subs:mcyt}

We considered the most common type of signature verification, namely deciding whether a test signature is a genuine signature or a skilled forgery (a signature imitated by a forger) in relation to a claimed identity (subject). To this end, we adopted the dataset MCYT-100~\cite{MCYT03} because it is one of the largest online signature datasets in the literature. MCTY-100 contains 100 subjects, each having 25 genuine signatures and 25 skilled forgeries. Each signature is a time series of 5D variables (2D coordinates, pressure, azimuth, and altitude angles of the pen tip). The temporal length ranges from a few hundred to a few thousand.


We tested the proposed approach with the following training protocol. The first 90/80/70/60/50\% of subjects in MCYT-100 were used for training and the remaining subjects for testing. Therefore, the training and test sets are completely non-overlapping in terms of both classes (subjects) and time series (signatures). During testing, for each subject in the test set, the first five genuine signatures were used as reference signatures. The remaining genuine signatures and all the skilled forgeries were used as test signatures. The distances between each test signature and all the reference signatures of the claimed subject were computed and their average was used as the distance between the test signature and the claimed identity. This distance can be passed to a subject-independent threshold classifier for the final decision of signature authenticity. {For each test signature, its distance to the corresponding five reference signatures is averaged. Based on this averaged distance, all test signatures (of all subjects) are then sorted and an Equal Error Rate (EER) is calculated, as in most related studies.}

\smallskip\noindent
\textbf{Implementation Details.} For MCYT-100, we adopted the deep dynamic time warping (DDTW)~\cite{WuKIUK19} for time series feature learning. Specifically, a Siamese network with two raw signatures as input and two feature sequences as output was trained by DDTW following the training protocol described above. The temporal length $W$ of each feature sequence is 256, and the number of dimensions $K$ is 64. This network was then used to extract feature sequences from all training, reference, and test signatures. Afterward, we built the proposed deep model with two feature sequences as input time series and a distance as output and trained it in the way described in Section~\ref{s:proposed}. During testing, this trained model was applied to the feature sequences of all reference and test signatures in order to calculate their distances. It is also possible to jointly learn feature representation and warping in an end-to-end manner, which will be our future work. Implementation details of DDTW can be found from its original paper~\cite{WuKIUK19}.

The U-net for MCYT-100 is the same as Unipen, but the encoder and decoder each have eight convolution layers. This difference is due to the different size of outer concatenation. The learning rate was set to 0.0001. The batch size was set to 15. The ratio of matching to non-matching pairs per batch was fixed at $1:2$.

\smallskip\noindent
\textbf{Baselines.} We compared our approach with the following four baselines.

\smallskip\noindent
\emph{DTW~\cite{Martinez-DiazFK14,WuKIUK19}.} In this baseline, each signature was represented by a handcrafted feature sequence~\cite{Martinez-DiazFK14}. The distances between test and reference signatures were then measured with DTW. In addition, the performance of DTW that takes raw signatures as inputs was also included in the comparison.

\smallskip\noindent
\emph{SN.} This SN has almost the same architecture as DDTW but was trained with conventional contrastive loss. Its output is not a feature sequence but a global feature vector produced by average pooling, and thus considers no temporal constraint. Another SN was also trained with the same local embedding loss as DDTW. It considers temporal constraint but offers limited temporal invariance.

\smallskip\noindent
\emph{DDTW~\cite{WuKIUK19}.} The only difference between DDTW and our approach lies in the warping process. DDTW uses DTW for warping, while our approach warps signatures with the proposed deep attention model.

\smallskip\noindent
\emph{Prewarping Siamese Network (PSN)~\cite{WuKUK19}.} This approach prewarps two signatures with DTW for temporal invariance. A Siamese network with the warped signatures as inputs and a distance as output is then trained with the same local embedding loss as DDTW. Similar to DDTW, PSN takes both temporal invariance and temporal constraint into account.

\begin{table}[t]
\caption{EER(\%) of online signature verification for MCYT-100.}
\label{t:MCYT}
\begin{tabular*}{\linewidth}{@{\extracolsep{\fill}}lccccc}
\toprule
& \multicolumn{5}{c}{Percentage of Training Data} \\
\cmidrule(lr){2-6}
~Method & 90\% & 80\% & 70\% & 60\% & 50\% \\
\midrule
~Proposed & \textbf{0.50} & \textbf{2.00} & \textbf{2.33} & \textbf{2.13} & \textbf{2.20} \\
\midrule
~DTW~\cite{Martinez-DiazFK14,WuKIUK19} & 4.00 & 3.00 & 4.17 & 4.37 & 4.60 \\
~~~w/ raw signatures as inputs & 5.00 & 6.25 & 5.73 & 6.37 & 6.96 \\
~SN & 5.50 & 6.80 & 6.27 & 7.33 & 8.40 \\
~~~w/ local embedding loss~\cite{WuKIUK19} & 3.50 & 3.40 & 3.75 & 3.75 & 5.50 \\
~DDTW~\cite{WuKIUK19} & 1.00 & 2.20 & 2.53 & 2.25 & 2.40 \\ ~PSN~\cite{WuKUK19} & \textbf{0.50} & 2.50 & 2.40 & 2.50 & 4.50 \\
\bottomrule
\end{tabular*}
\vspace{2mm}
\end{table}

\smallskip\noindent
\textbf{Results.} The EERs of our approach and the baselines are shown in Table.~\ref{t:MCYT}. In general, our approach outperformed all baselines in all experiments. Our lower EER than DTW indicates our superiority in the ability to learn complex representations over handcrafted features. Our approach also significantly outperformed its competitors, which only consider either temporal invariance (SN) or temporal constraints (SN w/ local embedding loss). Our improvement over DDTW shows the greater effectiveness of our trainable attention model in terms of warping and distance measurement compared to the traditional DTW.

\begin{figure}[t]
    \begin{minipage}{0.19\hsize}
    \begin{center}
    \includegraphics[width=0.95\linewidth]{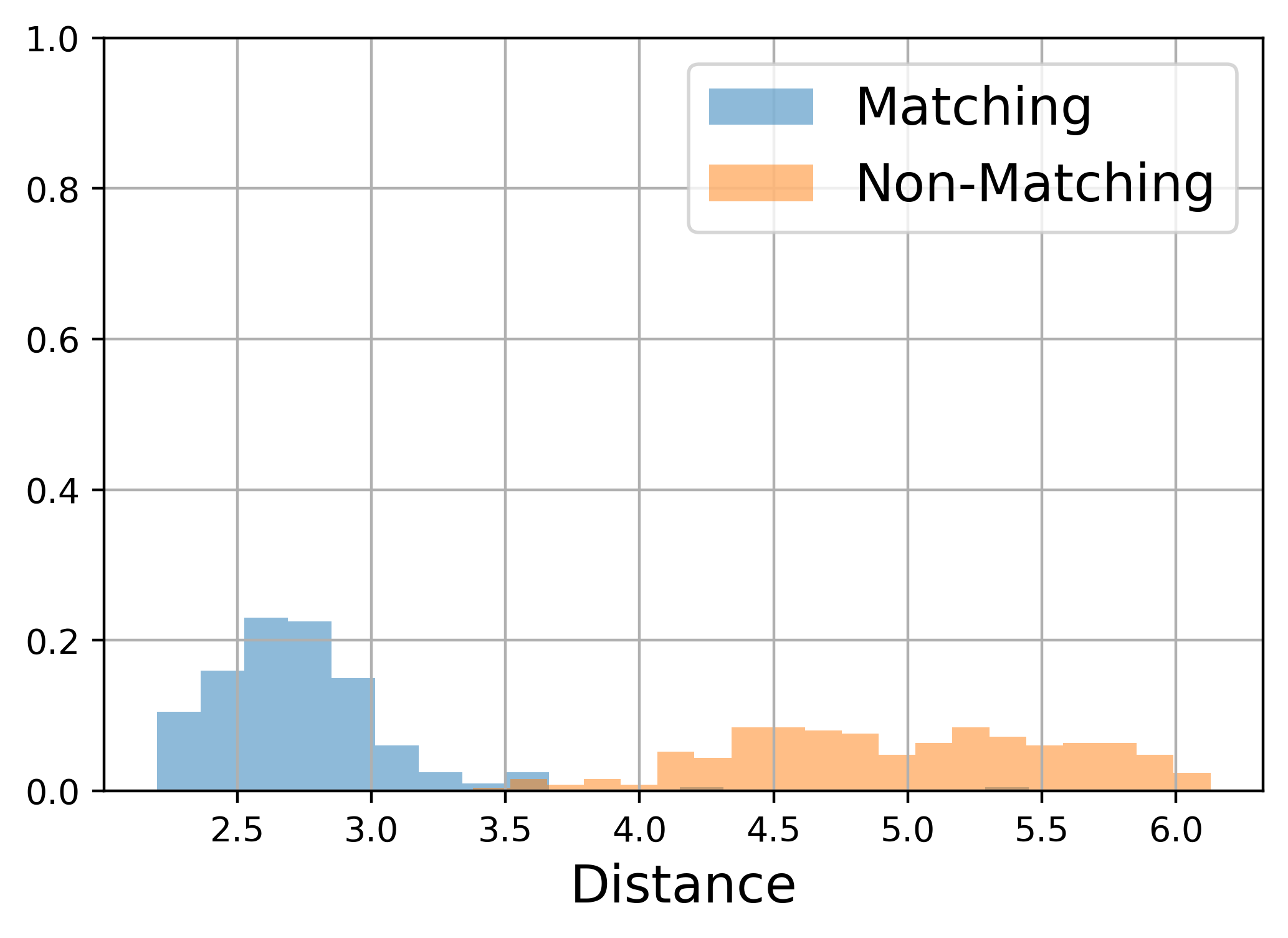}\\
    {\small (a)~90\%}
    \end{center}
    \end{minipage}
    \begin{minipage}{0.19\hsize}
    \begin{center}
    \includegraphics[width=0.95\linewidth]{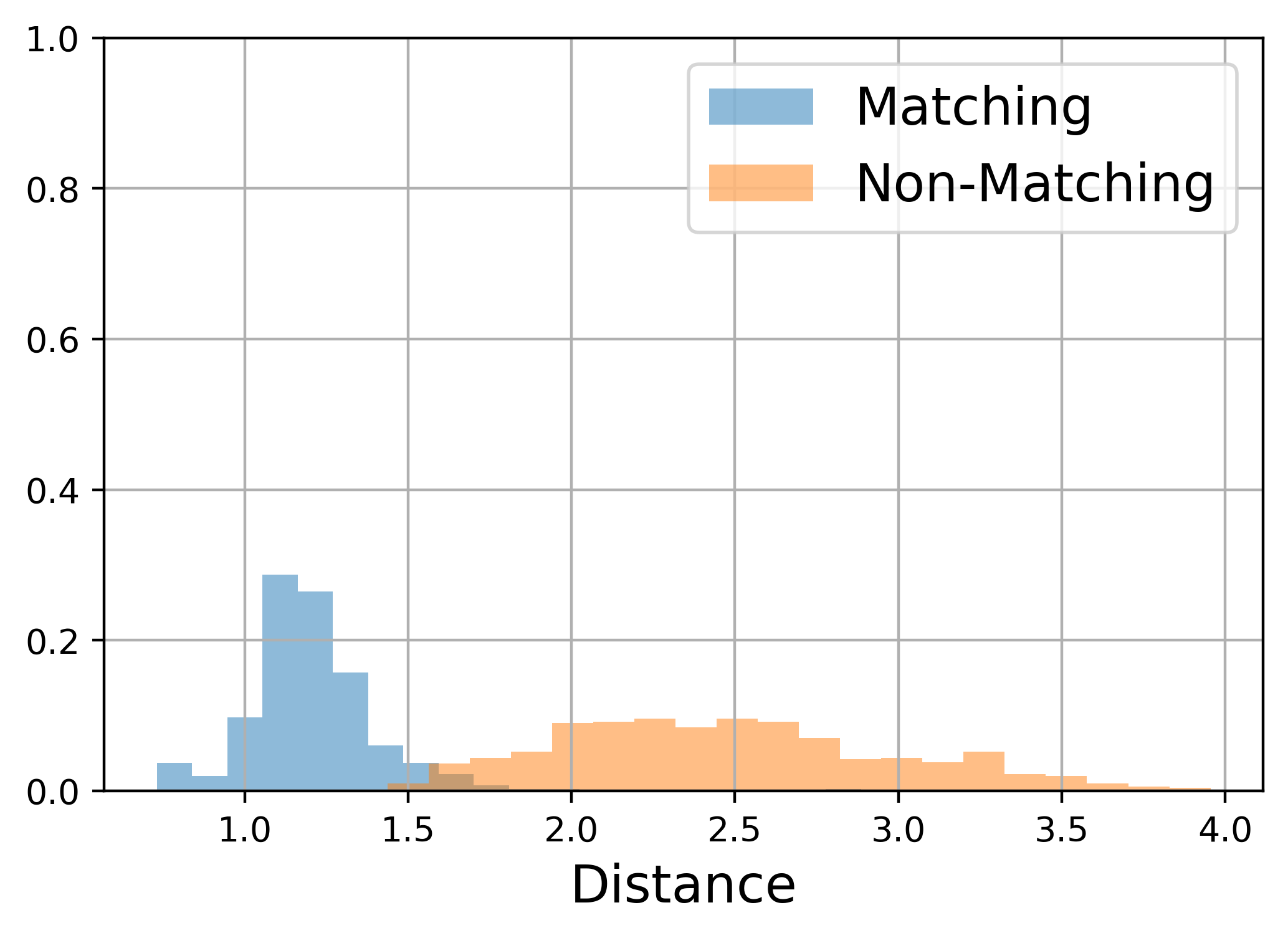}\\
    {\small (b)~80\%}
    \end{center}
    \end{minipage}
    \begin{minipage}{0.19\hsize}
    \begin{center}
    \includegraphics[width=0.95\linewidth]{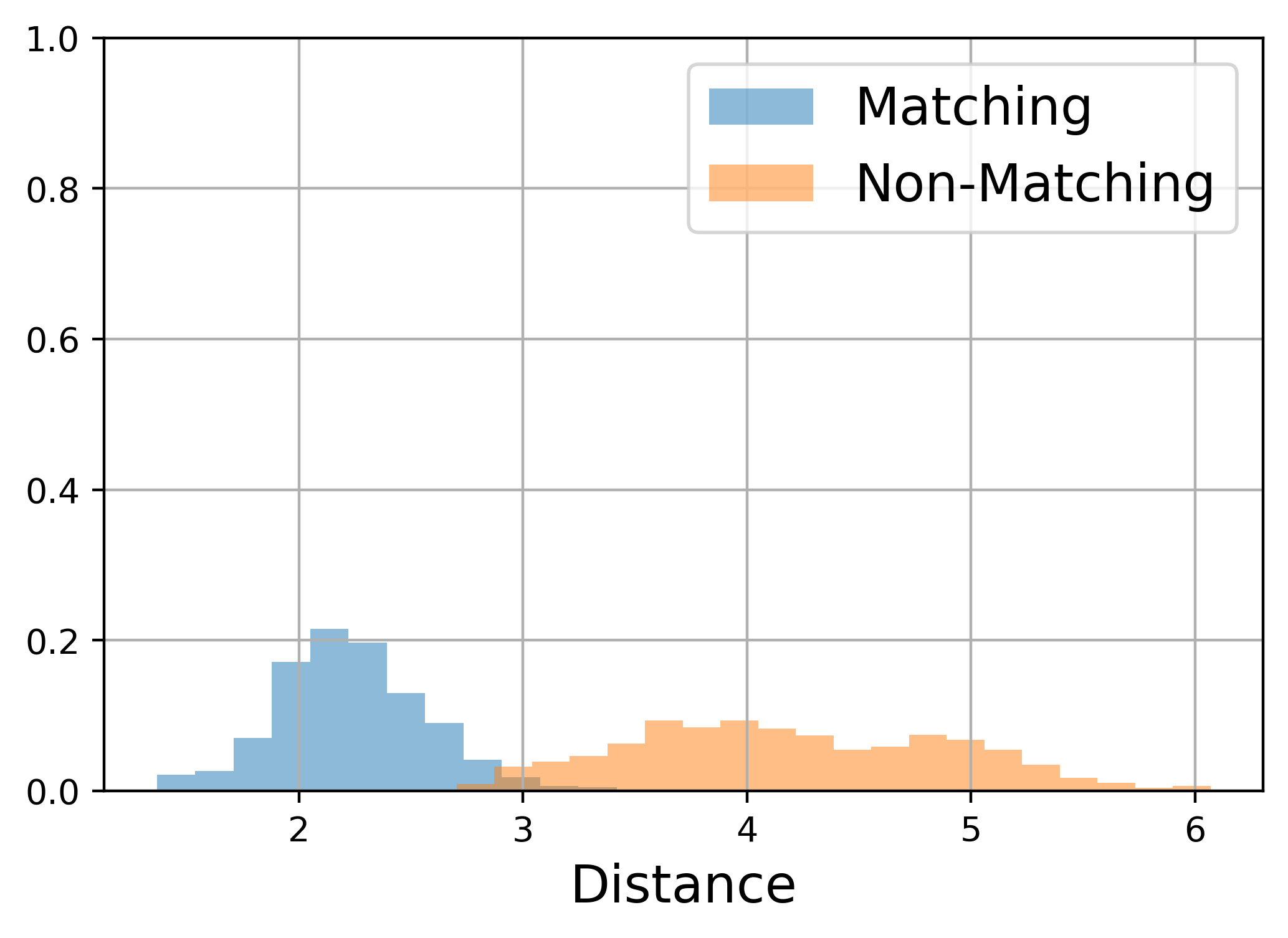}\\
    {\small (c)~70\%}
    \end{center}
    \end{minipage}
    \begin{minipage}{0.19\hsize}
    \begin{center}
    \includegraphics[width=0.95\linewidth]{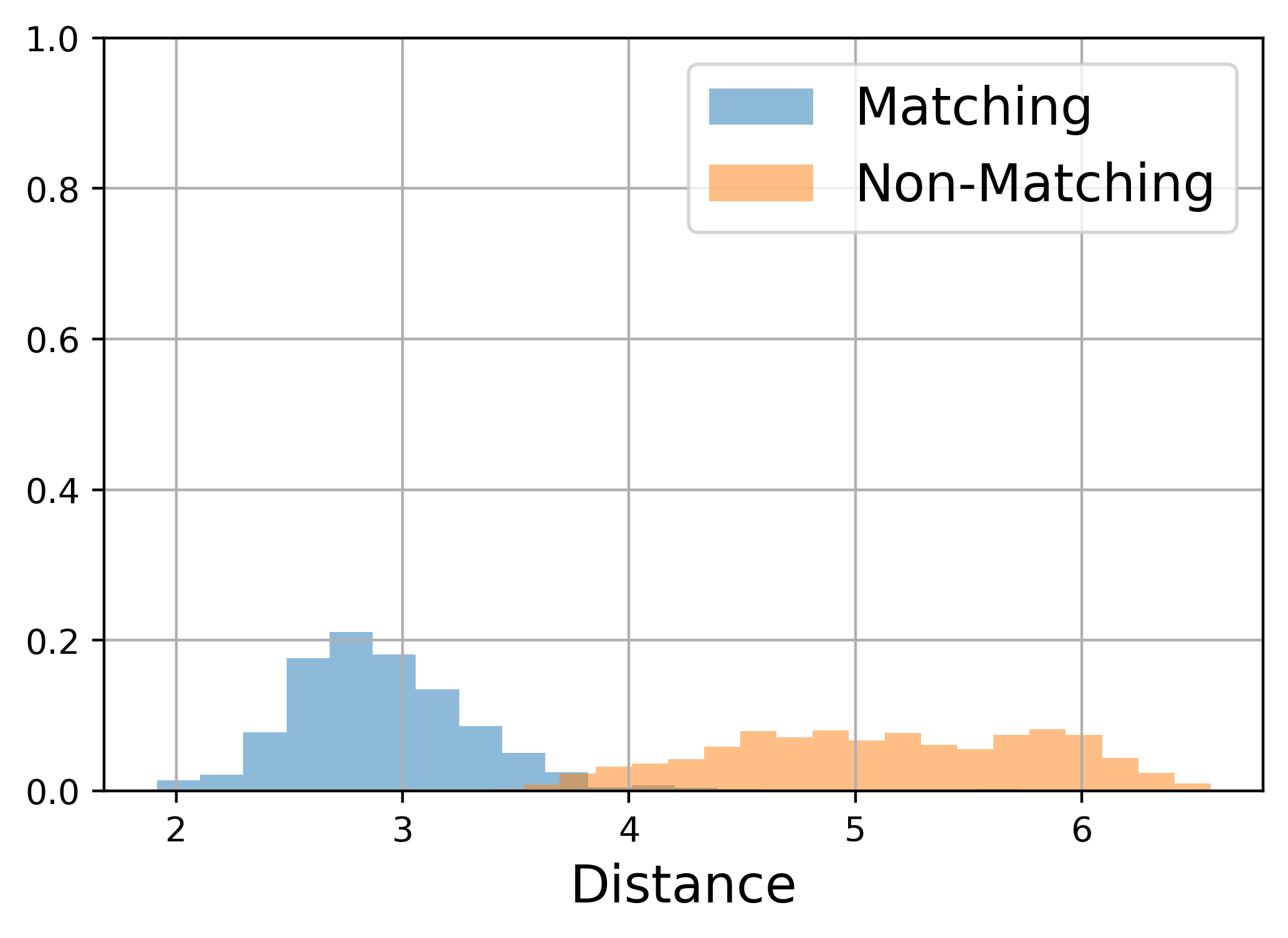}\\
    {\small (d)~60\%}
    \end{center}
    \end{minipage}
    \begin{minipage}{0.19\hsize}
    \begin{center}
    \includegraphics[width=0.95\linewidth]{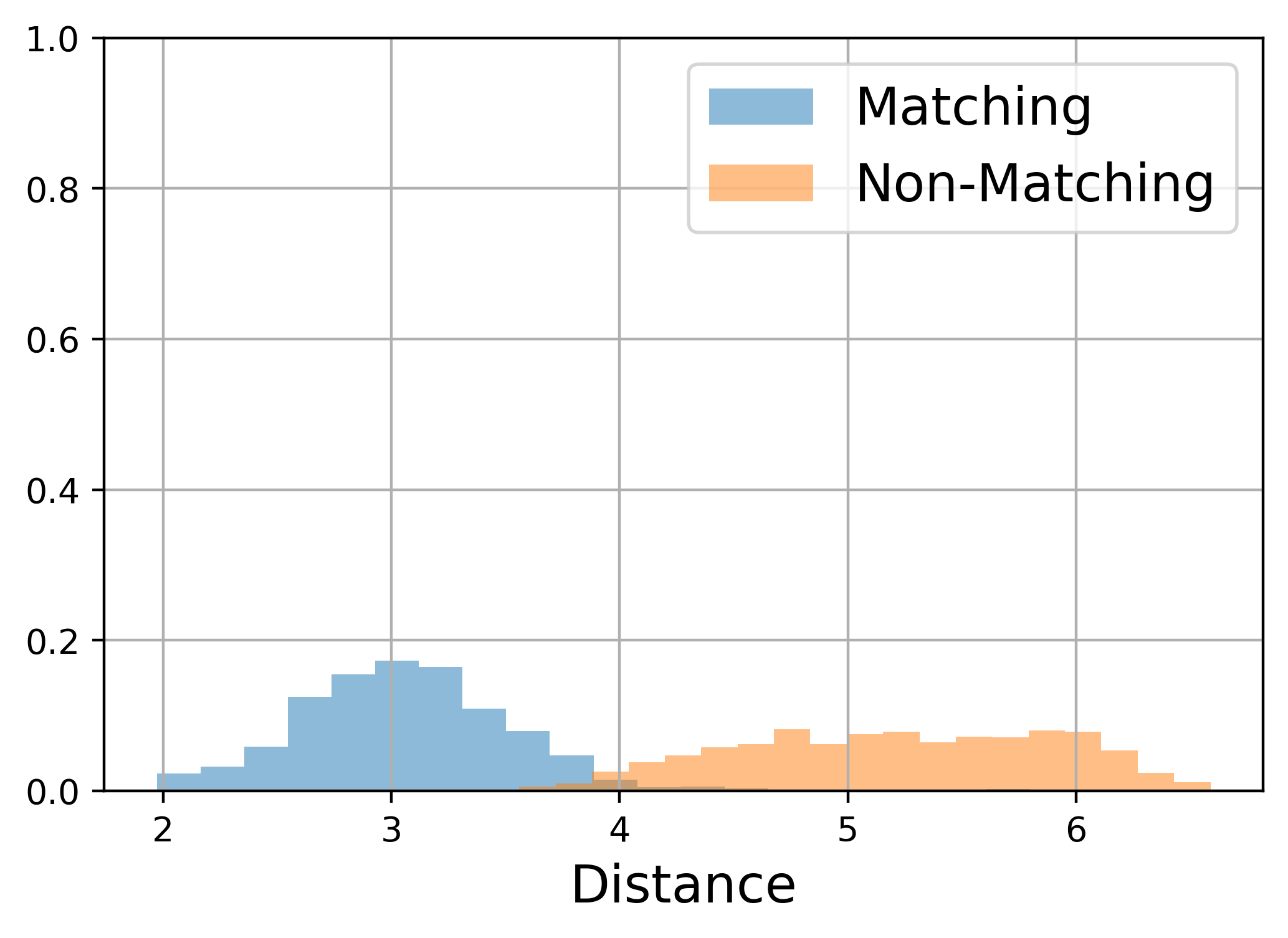}\\
    {\small (e)~50\%}
    \end{center}
    \end{minipage}
    \caption{Distributions of distances measured by DDTW~\cite{WuKIUK19} for MCYT-100. The sub-captions indicate the percentage of training data.}
    \label{fig:hist_dtw_signature}
    
    \vspace*{\floatsep}

    \begin{minipage}{0.19\hsize}
    \begin{center}
    \includegraphics[width=0.95\linewidth]{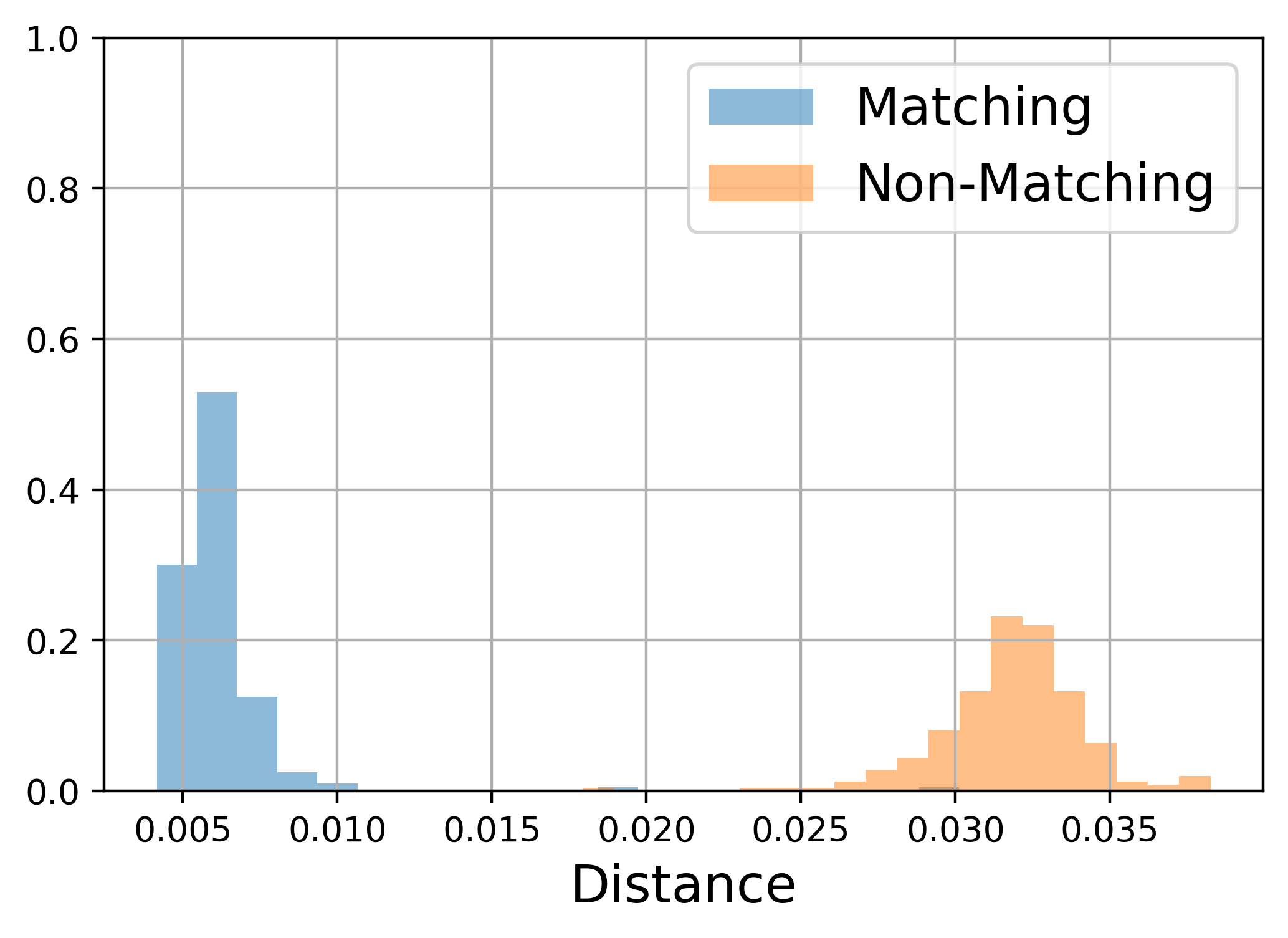}\\
    {\small (a)~90\%}
    \end{center}
    \end{minipage}
    \begin{minipage}{0.19\hsize}
    \begin{center}
    \includegraphics[width=0.95\linewidth]{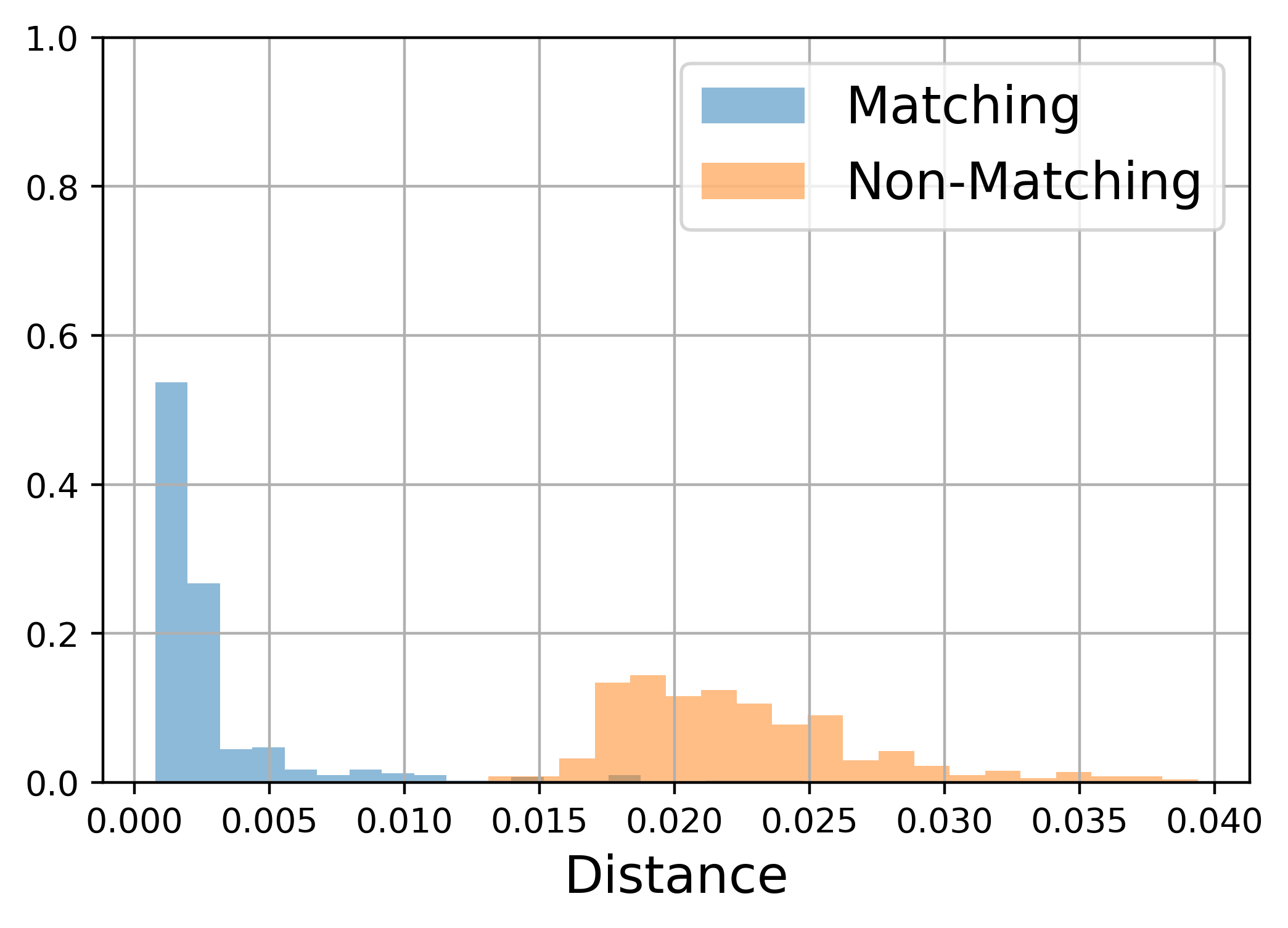}\\
    {\small (b)~80\%}
    \end{center}
    \end{minipage}
    \begin{minipage}{0.19\hsize}
    \begin{center}
    \includegraphics[width=0.95\linewidth]{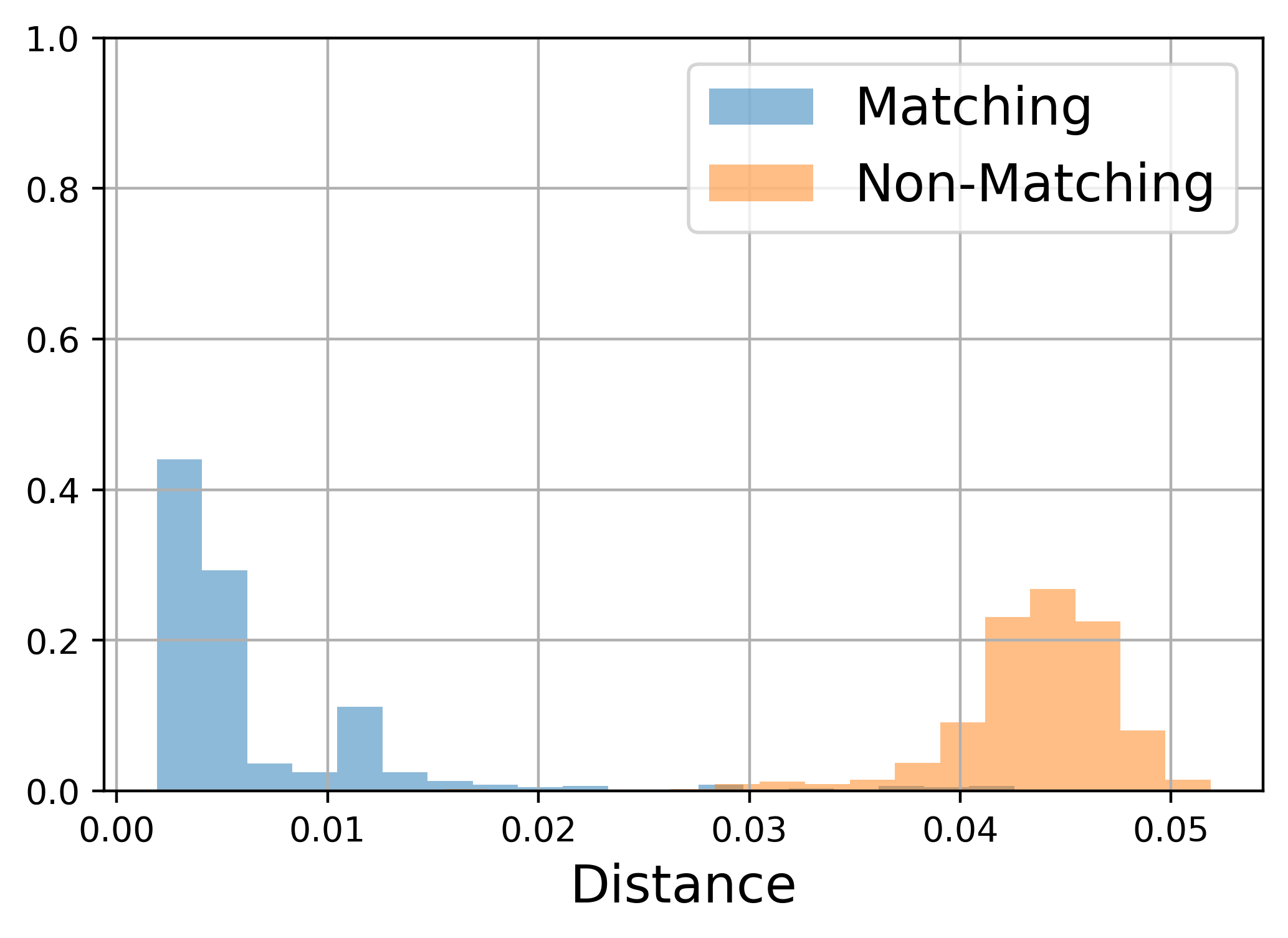}\\
    {\small (c)~70\%}
    \end{center}
    \end{minipage}
    \begin{minipage}{0.19\hsize}
    \begin{center}
    \includegraphics[width=0.95\linewidth]{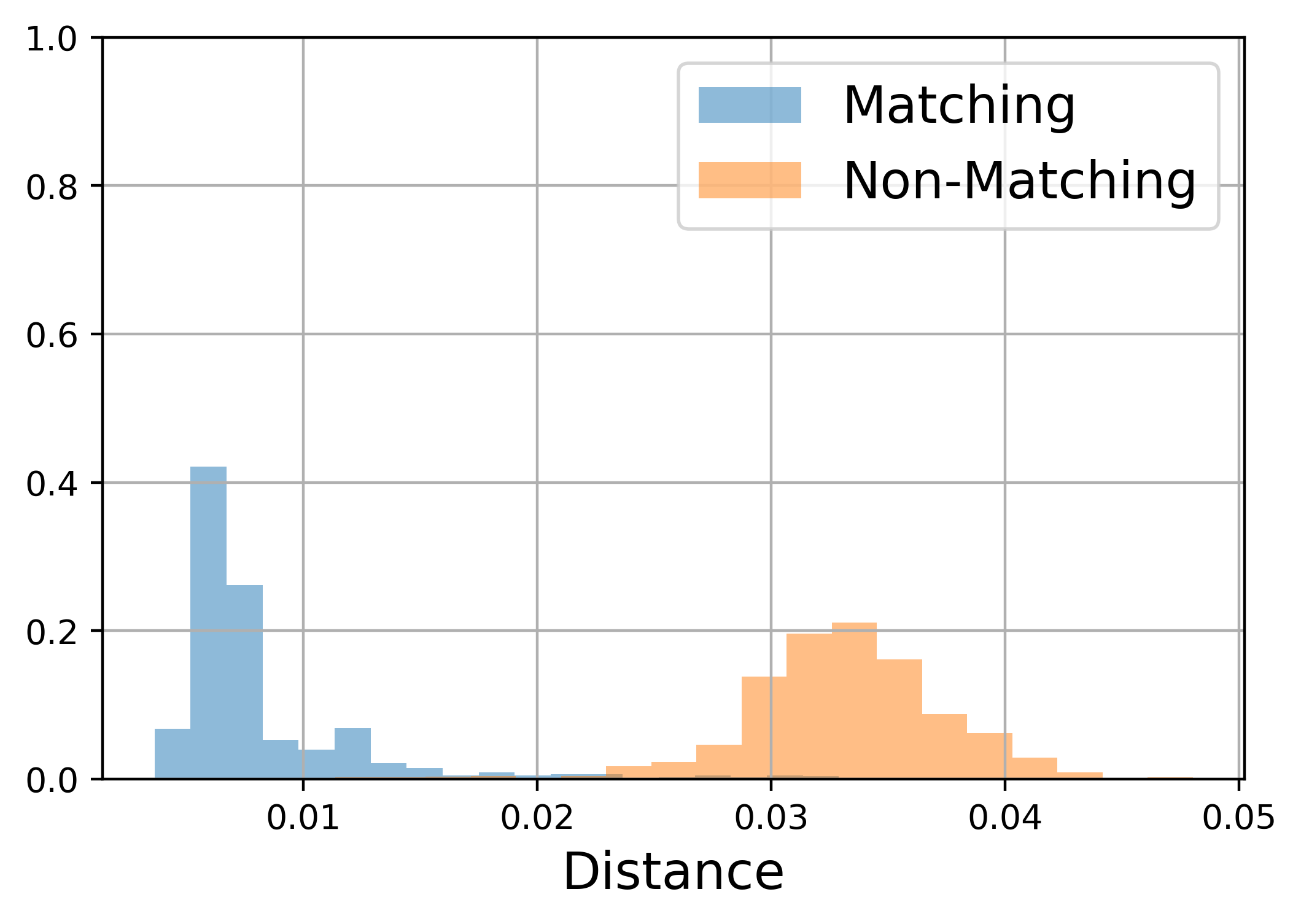}\\
    {\small (d)~60\%}
    \end{center}
    \end{minipage}
    \begin{minipage}{0.19\hsize}
    \begin{center}
    \includegraphics[width=0.95\linewidth]{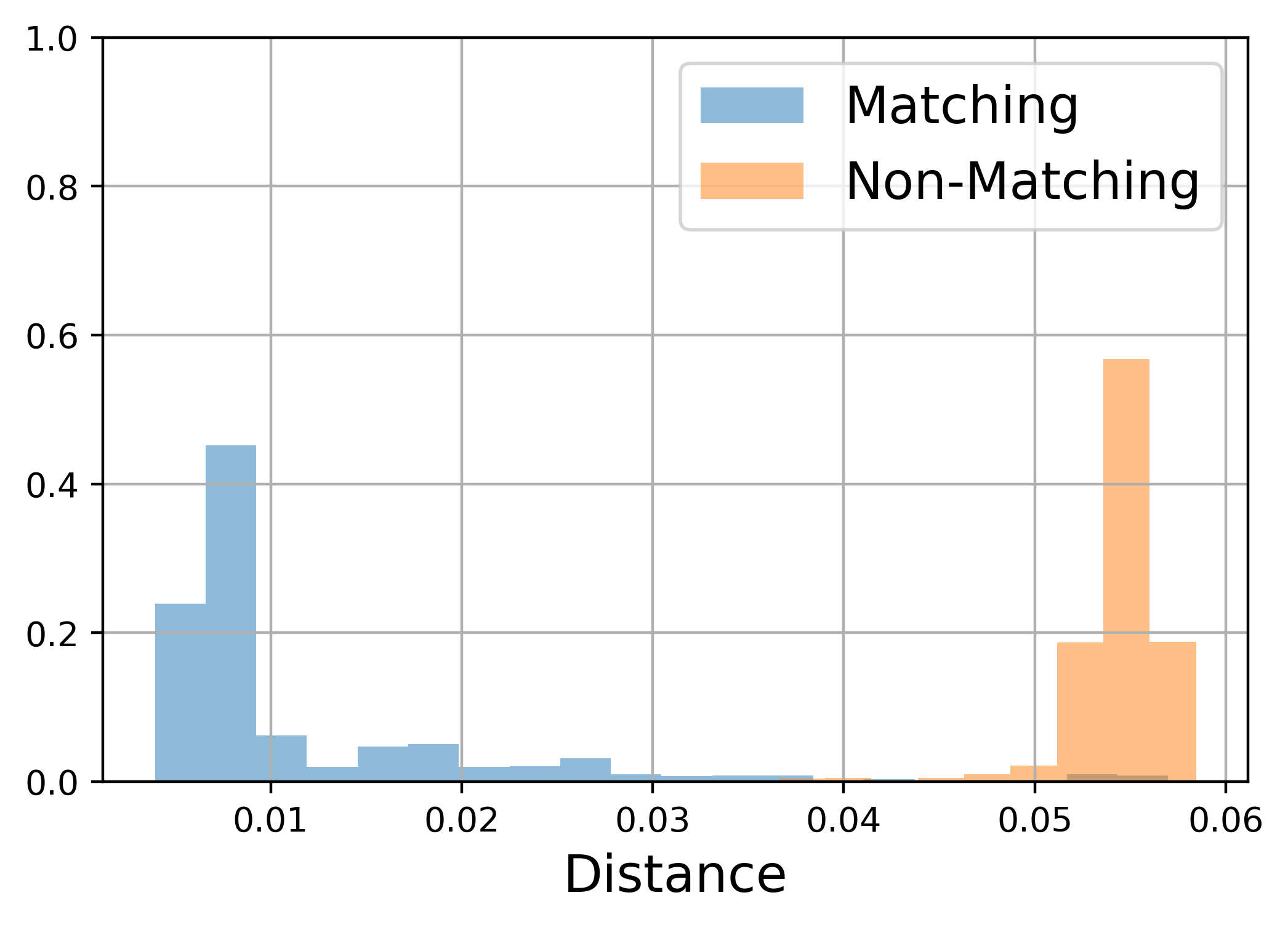}\\
    {\small (e)~50\%}
    \end{center}
    \end{minipage}
    \caption{Distributions of distances measured by proposed approach for MCYT-100. The sub-captions indicate the percentage of training data.}
    \label{fig:hist_proposed_signature}
\end{figure}

Figures~\ref{fig:hist_dtw_signature} and \ref{fig:hist_proposed_signature} show the histograms of the distances of matching and non-matching pairs, corresponding to DDTW and the proposed approach, respectively. We can see that our approach shows a much smaller overlap than DTW. As described above, DDTW focuses only on learning time series features, while our approach further improves warping through learning to facilitate easier separation of the distances of matching and non-matching pairs.

\begin{table}[t]
\caption{EERs (\%) published on MCYT. \#sub indicates the number of subjects in the test set. The number of reference signatures is five for all methods.}
\label{t:comparison-sota}
\centering
\begin{tabular*}{\linewidth}{@{\extracolsep{\fill}}lccc}
\toprule
~Method & \#sub & Deep? & EER (\%) \\
\midrule
~Yanikoglu \& Kholmatov~\cite{YanikogluK09} & 100 & \xmark & 7.80 \\
~Vivaracho{-}Pascual et al.~\cite{Vivaracho-PascualFP09} & 280 & \xmark & 6.60 \\
~Fa{\'{u}}ndez{-}Zanuy~\cite{Faundez-Zanuy07} & 280 & \xmark & 5.42 \\
~Nanni \& Lumini~\cite{NanniL08b} & 100 & \xmark & 5.20 \\
~Cpalka et al.~\cite{CpalkaZR16} & 100 & \xmark & 4.88 \\
~Sae{-}Bae \& Memon~\cite{Sae-BaeM14} & 100 & \xmark & 4.02 \\
~Tang et al.~\cite{TangKF18} & 100 & \xmark & 3.16 \\
\midrule
~DDTW~\cite{WuKIUK19} & 50 & \cmark & 2.40 \\
~PSN~\cite{WuKUK19} & 50 & \cmark & 4.50 \\
\midrule
~Proposed Approach & 50 & \cmark & \textbf{2.20} \\
\bottomrule
\end{tabular*}
\vspace{2mm}
\end{table}

We also tested our approach without pre-training, where He initialization~\cite{HeZRS15} was used to initialize the model weights. However, the contrastive losses could not decrease as the training progresses. The reason may be due to the difficulty of the signature verification task, such as the small amount of training data and the subtle difference between genuine signatures and skilled forgeries. The failure of He initialization confirms the necessity of pre-training based on DTW.


Table~\ref{t:comparison-sota} compares our performance with the state-of-the-art EERs previously reported for MCYT. {Noted that similar to DDTW and PSN, our model is learning-based and so our superior accuracy was obtained using only the test set, which contains many fewer subjects. The comparison is thus biased toward our model relative to the state of the art.} {In the future, we shall consider training the network with a larger dataset and testing it on the whole of MCYT-100.}


{After training with 90\% of the subjects, the EER of our approach was 0.5\%, as shown in Table~\ref{t:MCYT}. Using similar experimental protocols, the EER of SynSig2Vec~\cite{LaiJLZM20} was 1.7\%, the EER of Li et al.~\cite{LiZLWLZW19} was 10.5\%, and the accuracy of OSVNet~\cite{VoruguntiSMP19} was 93.0\%. All of these previous studies are deep signature verification approaches. These results demonstrate the superiority of our approach over the state of the art. In this section, we have applied our approach only to skilled forgery verification, but it can be easily adapted to identify random forgeries. We will demonstrate this in the future. We also plan to test our approach on a larger dataset such as DeepSignDB~\cite{TolosanaVFO21}.}

\subsection{Qualitative Analysis}

\begin{figure}[t]
\centering

\subfloat[Matching pairs]{
\includegraphics[width=.7\linewidth]{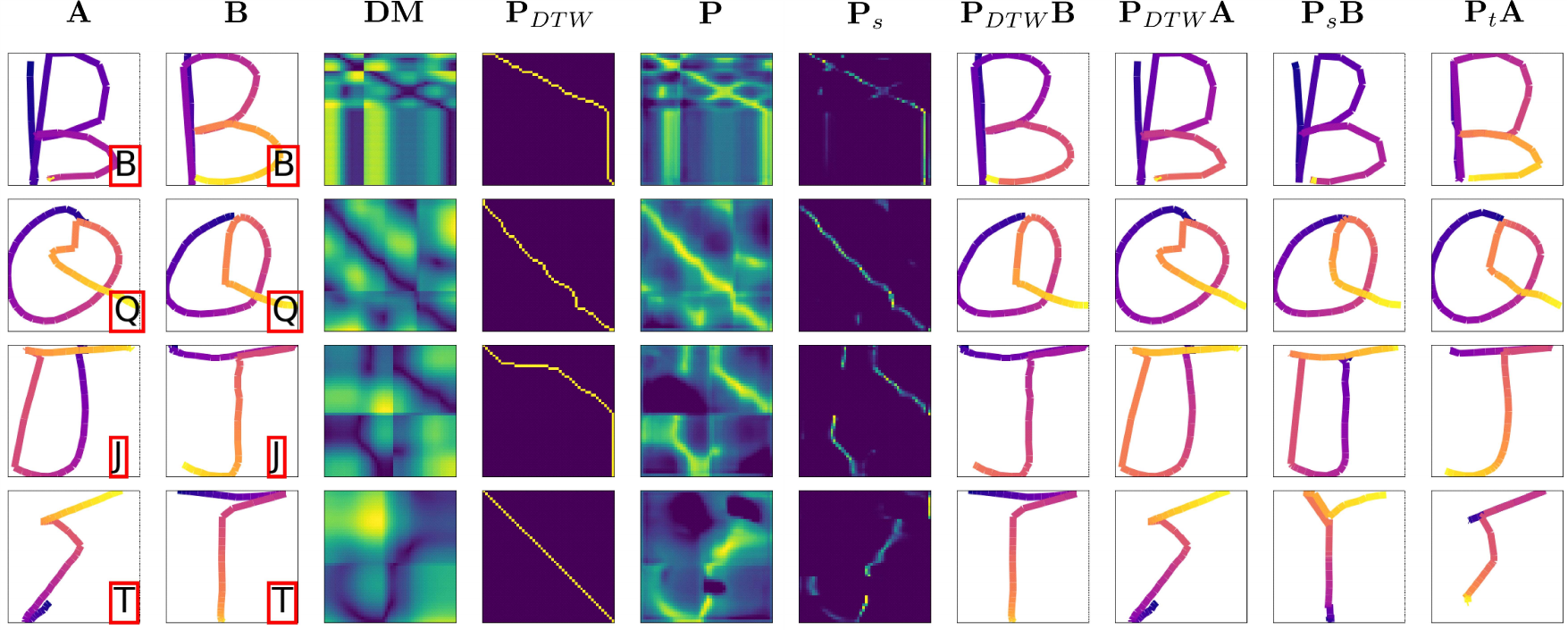}
\label{subf:matching_pair}}
\\
\subfloat[Non-matching pairs]{
\includegraphics[width=.7\linewidth]{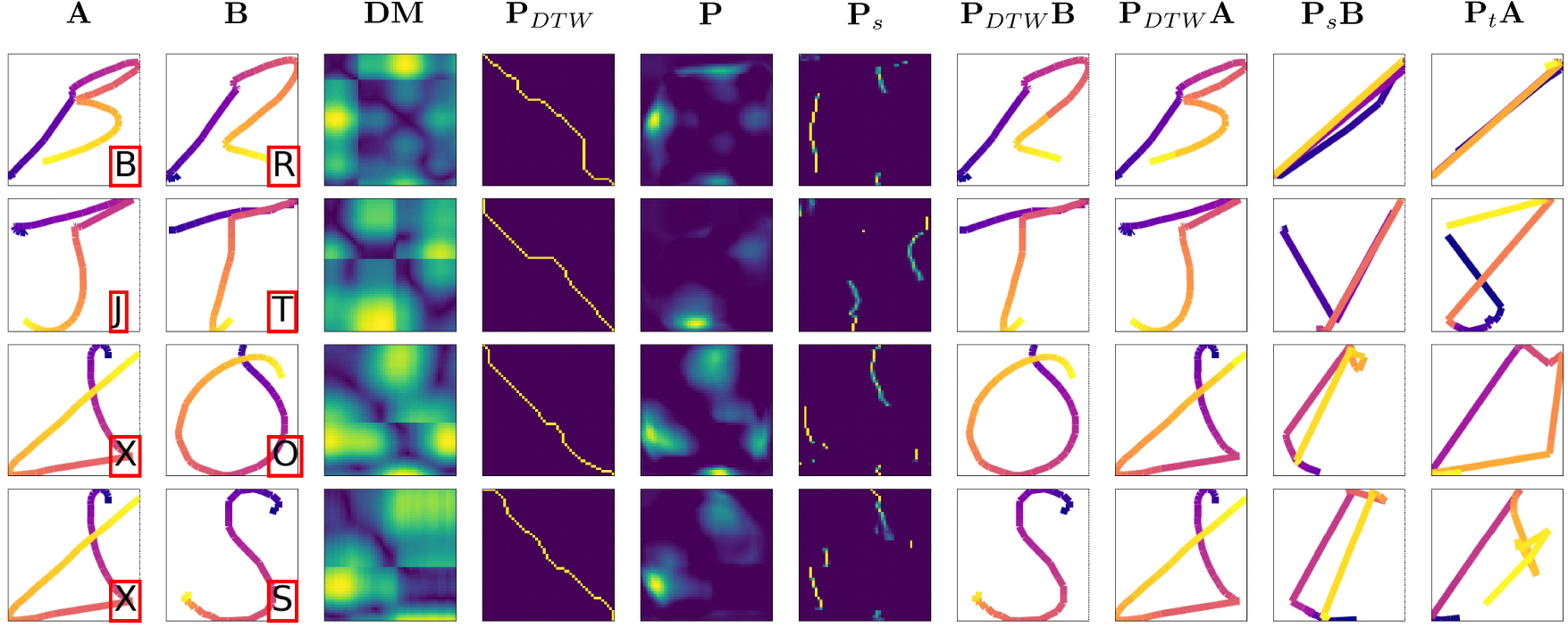}
\label{subf:non-matching_pair}}
\caption{Examples of input time series $\mathbf{A}$ and $\mathbf{B}$, DTW distance matrix (DM), DTW warping path $\mathbf{P}_\mathrm{DTW}$, output $\mathbf{P}$ of U-net, our warping matrix $\mathbf{P}_s$, DTW-warped $\mathbf{B}$ and $\mathbf{A}$, and our warped time series $\mathbf{P}_s\mathbf{B}$ and $\mathbf{P}_t\mathbf{A}$. Each time series starts with the darkest color. For DM, $\mathbf{P}_\mathrm{DTW}$, $\mathbf{P}$, and $\mathbf{P}_s$, the brighter the color, the larger the value.}
\label{f:visualization_pair}
\end{figure}


\begin{figure}[t]
\centering
\includegraphics[width=.7\linewidth]{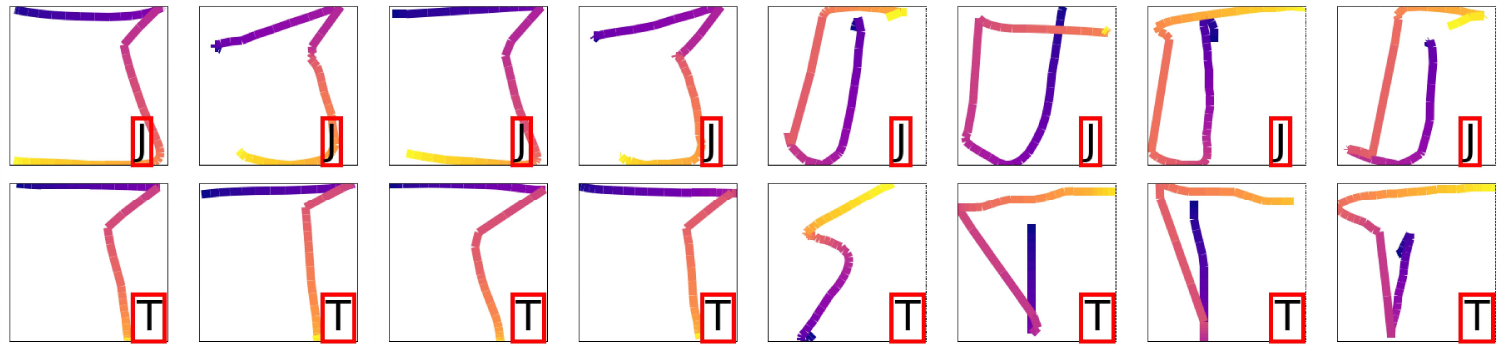}
\caption{Examples of Unipen training data of \texttt{J} (top) and \texttt{T} (bottom). Each time series starts with the darkest color and finishes with the lightest color. For both classes, the leftmost four letters have a totally different stroke order from the rightmost ones.}
\label{f:training_data}
\end{figure}

Figure~\ref{f:visualization_pair} shows examples of warping behavior of our approach on the Unipen dataset. In all examples, our network correctly classified the two time series into matching and non-matching pairs. Some interesting behaviors can be observed. For relatively easy matching pairs (first two rows in Fig.~\ref{subf:matching_pair}), our network predicted warp paths that satisfied the monotonicity, continuity, and boundary conditions similar to DTW. When two letters of the same class but with completely different stroke orders came along (last two rows in Fig.~\ref{subf:matching_pair}), our network could learn to adaptively break through the temporal constraints described above, defying the common wisdom of time warping. The correct classifications shown in the last two examples in Fig.~\ref{subf:matching_pair} are not accidental, but because the training data includes such intra-class variation in stroke order (Fig.~\ref{f:training_data}) and thanks to the great robustness of our attention model. The conventional DTW is obviously not capable of handling such challenging time series.

Figure~\ref{subf:non-matching_pair} shows some challenging examples of non-matching pairs that are very similar in both appearance and stroke order. The warping suggested by DTW could not effectively distinguish these pairs. In comparison, if we focus on the pair of input handwriting $\mathbf{A}$ vs. warped \say{handwriting} $\mathbf{P}_s\mathbf{B}$ (or $\mathbf{B}$ vs. $\mathbf{P}_t\mathbf{A}$), we can observe a significant difference in appearance. That is, our network could learn to augment the inter-class variation through warping, so that similar but different classes could be distinguished in a much easier manner. These results demonstrate the high discriminative power of our approach.



\section{Conclusion}


We proposed to learn a warping function to align the indices of time series using an attention model specialized for metric learning. This model is invariant not only to local or semi-local temporal distortions, but also to large global ones, so that even matching pairs that do not satisfy the monotonicity, continuity, and boundary conditions can still be successfully identified. We also proposed to pre-train our model with the guidance of DTW for stabilized learning and higher discriminative power.

The proposed pre-training can actually be accomplished using general time series data that do not have to be included in the training set. If this is true, our model would be relieved of the requirement for large-scale labeled data in the target (classification or verification) domain. This property would be beneficial for tasks, e.g., signature verification or writer identification, where only a small dataset is available. We will investigate this in future research.




%
%
%

\bibliographystyle{splncs04}
\bibliography{refs}
\end{document}